\documentclass[runningheads]{llncs}
\usepackage[T1]{fontenc}
\usepackage{cite}
\usepackage{amsmath}
\usepackage{graphicx}
\usepackage{multirow}
\usepackage{textcomp,comment}
\usepackage{xcolor}
\usepackage{multirow}
\usepackage{booktabs}
\usepackage{url}
\usepackage{amsfonts}
\usepackage{marvosym}
\begin{document}

\title{DSTFView: Multi-View Cloud-Edge Workload Forecasting with Dual-Input Spatio-Temporal-Frequency Modeling}
\titlerunning{DSTFView for Cloud-Edge Workload Forecasting}

\author{Qingzhong Li$^1$,  Hui Ma$^1$(\Letter), Yajun Zhang$^1$, Qingchang Ma$^1$, Zhou Long$^1$}

\authorrunning{Q. Li, H. Ma, and Y. Zhang, et al.}

\institute{
$^1$Xinjiang Key Laboratory of Intelligent Computing and Smart Applications, School of Software,
Xinjiang University, Urumqi, 830017, China \\
\email{liqingzhong@stu.xju.edu.cn, \{huima, zyj\}@xju.edu.cn, \{maqingchang, longzhou\}@stu.xju.edu.cn}}

\maketitle     
\begin{abstract} 
With the widespread deployment of edge-side AI inference, edge platforms are increasingly required to support latency-sensitive, highly concurrent, and reliability-critical applications. However, existing methods often struggle to balance multidimensional feature modeling and forecasting efficiency in collaborative cloud-edge environments. To address this issue, we propose DSTFView, a dual-input spatio-temporal-frequency multi-view workload forecasting framework for collaborative cloud-edge environments. It jointly models closeness and period dependencies and extracts spatial, temporal, and frequency-domain dependencies. Besides, it designs an adaptive fusion mechanism and adjusts the contribution of each view  to capture abrupt changes. Experimental results on the CPU and TP datasets demonstrate that DSTFView consistently outperforms representative baselines across multiple forecasting horizons and evaluation metrics.

\end{abstract}
\section{Introduction}
With the ongoing evolution of edge-cloud infrastructures and the rapid expansion of edge-side AI inference services, edge platforms are increasingly expected to accommodate latency-sensitive and highly concurrent applications despite limited resource availability~\cite{ning2023uav_mec_ml_survey,zhu2023pushing}. Under such circumstances, edge workload forecasting is important for optimizing resource management, decreasing resource waste and operational expenses while preserving service quality~\cite{ghahari2025efficient}.

However, edge workload forecasting remains a challenging task because it is difficult to uniformly characterize and integrate the heterogeneous, multi-source, and multi-pattern features in edge workloads~\cite{Ma2024EdgeWorkloadPrediction}. Specifically, edge workloads typically exhibit spatial dependencies across sites or virtual machines, temporal dynamics such as burstiness and trend shifts, and frequency-domain regularities arising from service cycles and recurring usage behaviors. Meanwhile, edge environments also involve multi-relational and time-varying interactions, including geographic proximity, service affinity, and shared-resource coupling. However, due to dynamic deployment, user mobility, and other latent factors, these dependency relationships are often only partially observable and inherently uncertain.
As a result, existing methods that focus on temporal dependencies~\cite{Chen2024Pathformer,wu2023timesnet,liu2023itransformer,zhang2023crossformer}, frequency-aware patterns~\cite{Fan2022DEPTS}, or spatio-temporal dependencies~\cite{Wang2025STGraphormer} from a single or local perspective are difficult to directly apply to edge workload forecasting. This issue becomes particularly critical in multi-step forecasting, where the lack of unified modeling for heterogeneous relations and multi-dimensional features can easily lead to poor prediction performance.

Another important challenge lies in balancing prediction accuracy and system efficiency in collaborative cloud-edge forecasting~\cite{Nrekovic2025Reducing}. Methods that rely only on edge-side information may overlook cross-site dependencies, whereas cloud-centric solutions often introduce noticeable uplink delay and communication cost. To alleviate this issue, collaborative cloud-edge paradigms have been investigated by combining cloud-side global modeling with edge-side local refinement~\cite{li2023elastic}. 
However, existing cloud-edge frameworks perform poorly in multi-step forecasting, especially when the exchanged representations cannot adequately capture multi-scale periodic behaviors, or when heterogeneous relations bring in task-irrelevant or unreliable connections that weaken correlation modeling over extended horizons.

To address the aforementioned challenges, we propose DSTFView, an edge workload forecasting framework for collaborative cloud-edge architectures. DSTFView follows a two-stage collaborative cloud-edge forecasting pipeline and introduces a dual-input spatio-temporal-frequency modeling mechanism into both the global and local stages, where the closeness input captures recent dynamics and the period input models recurring temporal patterns. At the cloud-side global stage, the framework leverages lightweight aggregated signals to learn site-level global priors, thereby exploring cross-site dependencies while controlling communication overhead. At the edge-side local stage, it further integrates fine-grained workload sequences with cloud-provided global guidance to refine VM-level multi-step forecasts. In addition, DSTFView incorporates an adaptive multi-view graph fusion mechanism into both stages, enabling the model to dynamically integrate different relational views and enhance dependency modeling.

The main contributions of this work are summarized as follows. \textit{Firstly}, we propose a novel dual-input spatio-temporal-frequency architecture for collaborative cloud-edge multi-step forecasting. \textit{Secondly}, we design an adaptive multi-view graph fusion strategy to dynamically integrate different relational views under varying workload patterns. \textit{Thirdly}, we build a global-local collaborative forecasting pipeline and conduct extensive experiments on the CPU and TP datasets. The results demonstrate that DSTFView consistently outperforms baseline methods.

\section{Related Work}
\subsection{Time Series Prediction}
Recently, cloud-side time series forecasting can generally be categorized into three main approaches: temporal, spatio-temporal, and frequency-aware forecasting methods~\cite{Kong2025DLTSFSurvey}.

\textit{Temporal forecasting methods.} Nie et al.~\cite{Yuqietal-2023-PatchTST} introduced PatchTST, which divides time series into patches and learns temporal dependencies with a Transformer-based architecture. Wu et al.~\cite{wu2023timesnet} proposed TimesNet, which explicitly models multi-period patterns in long-range forecasting. Liu et al.~\cite{liu2023itransformer} developed iTransformer by reformulating the attention mechanism to capture variable-wise interactions, whereas Zhang et al.~\cite{zhang2023crossformer} presented Crossformer to model cross-scale dependencies through hierarchical feature interactions.

\textit{Spatio-temporal forecasting methods.} Xie et al.~\cite{xie2023megacrn} proposed MegaCRN, which incorporates memory-based components to capture dynamic interactions among nodes. Zhou et al.~\cite{zhou2024stgormer} developed a graph-transformer-based forecasting model that injects spatio-temporal structural information into the attention mechanism and employs specialized modules to characterize heterogeneous traffic dynamics. Zheng and Xie~\cite{zheng2025dstsgnn} introduced DST-SGNN, which combines Stiefel-manifold-constrained spectral graph operations with an efficient dynamic graph optimization strategy to learn time-varying spatial relationships.

\textit{Frequency-aware forecasting methods.} They emphasize spectral representations for capturing periodicity and multi-scale patterns. Zhou et al.~\cite{zhou2022fedformer} proposed FEDformer, which utilizes frequency-domain operators to extract informative spectral components for long-horizon prediction. Zhao et al.~\cite{zhao2024tfegru} introduced TFEGRU, which incorporates a time-frequency enhancement module and combines GRU with multi-head self-attention for cloud workload forecasting. 

However, existing methods cannot effectively model heterogeneous multi-source relations and spatio-temporal-frequency features, which leads to poor performance in edge workload forecasting.

\subsection{Collaborative Cloud–Edge Forecasting}
Collaborative forecasting in cloud--edge environments seeks to balance prediction accuracy and system cost by integrating cloud-level global modeling with edge-level local adaptation. Li et al.~\cite{li2023elastic} proposed ELASTIC, a two-stage framework that aggregates VM-level signals at the edge for cloud-side global modeling and then refines predictions through edge-side disaggregation. Also, they~\cite{10759304} extended this idea with improved aggregation/disaggregation modules and extra training objectives for stronger cloud--edge collaboration.

However, existing methods often capture only limited inter-site relations or use coarse fusion strategies, which makes it difficult to model the complex dependencies in edge environments, particularly when workload periodicity is strong and site correlations are diverse. 

\section{Method}
\subsection{Framework Overview}

\begin{figure*}[t]
  \centering
  \includegraphics[width=0.95\linewidth]{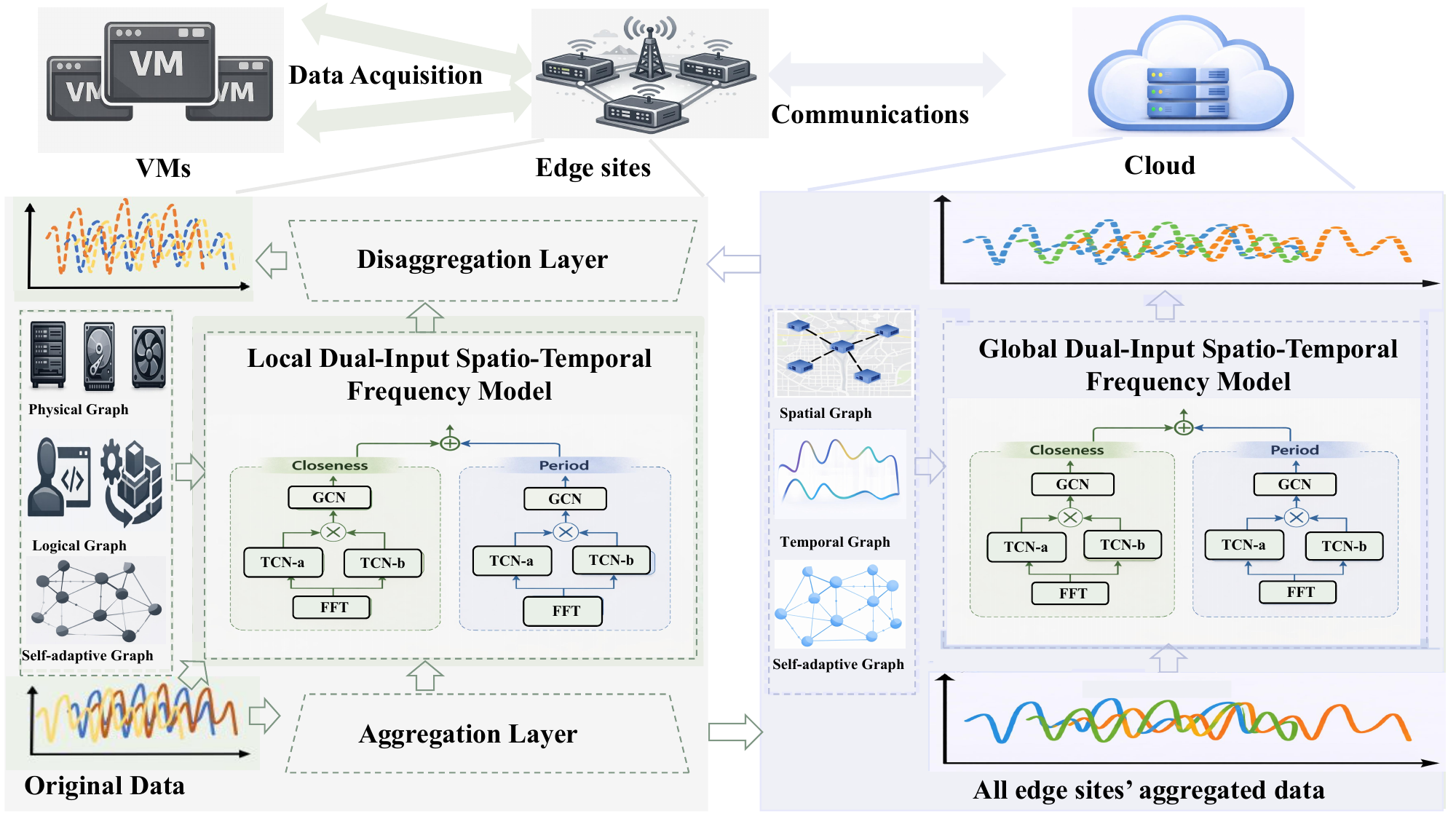}
  \caption{Structure of DSTFView model.}
  \label{fig:framework}
\end{figure*}
As shown in Fig.~\ref{fig:framework}, \textsc{DSTFView} adopts a collaborative two-stage cloud-edge forecasting framework. At the edge side, historical VM-level workload traces are collected at each site and compressed by a learnable \emph{Aggregation Layer} into lightweight site-level sequences. These aggregated representations are then transmitted to the cloud to reduce communication cost. In the cloud-side global stage, the uploaded sequences from all participating sites are used to perform global dual-input spatio-temporal-frequency (STF) modeling over multi-view site graphs, producing coarse-grained site-level multi-step forecasts. The cloud-side predictions are subsequently returned to each edge site, where a local model carries out dual-input STF modeling at VM granularity based on intra-site multi-view graphs. Finally, a \emph{Disaggregation Layer} projects the site-level cloud forecasts back to the VM level and combines them with local predictions, yielding the final VM-level multi-step forecasting results.
\subsection{Notations and Problem Formulation}
We consider a collaborative cloud--edge paradigm with $M$ edge sites, where site $m$ hosts $N_m$ VMs. Let $C$ denote the number of workload variables and $H$ the prediction horizon. For dual-input temporal modeling, let $P_c$ and $P_p$ denote the lengths of the closeness sequence and the period sequence, respectively.

At time $t$, the VM-level closeness input and period input at site $m$ are denoted by $\mathbf{X}_m^c\in\mathbb{R}^{N_m\times P_c\times C}$ and $\mathbf{X}_m^p\in\mathbb{R}^{N_m\times P_p\times C}$, respectively. The prediction target is denoted by $\mathbf{X}_m^{t:t+H-1}\in\mathbb{R}^{N_m\times H\times C}$. Our goal is to learn a collaborative forecaster $\mathcal{F}$ such that
\begin{equation}
\widehat{\mathbf{X}}_m^{t:t+H-1}
=
\mathcal{F}\!\left(\mathbf{X}_m^c,\mathbf{X}_m^p\right),
\quad
\forall m\in\{1,\dots,M\}.
\end{equation}

\noindent\textbf{Notation convention.}
Scalars are denoted by lowercase symbols (e.g., $P_c,P_p,H,C$); vectors by bold lowercase symbols; matrices/tensors by bold uppercase symbols (e.g., $\mathbf{X},\mathbf{A}$); and functions/operators by symbols (e.g., $\mathcal{F}, f_{\mathrm{dis}}, \phi$).

\subsection{Global Stage: Aggregation and Cloud-side STF Forecasting}
The global stage consists of an aggregation layer deployed at each edge site and a cloud-side dual-input STF predictor. Its goal is to compress VM-level traces into lightweight site-level representations and capture cross-site dependencies in the cloud.

For each site $m$, the VM-level closeness and period inputs are separately compressed into site-level representations through a learnable aggregation layer, yielding $\mathbf{Y}_m^c$ and $\mathbf{Y}_m^p$, respectively, where $\mathbf{Y}_m^c\in\mathbb{R}^{1\times P_c\times C}$ and $\mathbf{Y}_m^p\in\mathbb{R}^{1\times P_p\times C}$. By stacking the outputs of all sites, we obtain the cloud-side dual inputs $\mathbf{Y}^c\in\mathbb{R}^{M\times P_c\times C}$ and $\mathbf{Y}^p\in\mathbb{R}^{M\times P_p\times C}$.

On the cloud side, we construct an inter-site multi-view graph set, denoted as $\mathcal{G}_g=[{\mathbf{A}_g^{(s)}, \mathbf{A}_g^{(t)}, \mathbf{A}_g^{(a)}}]$, where $\mathbf{A}_g^{(s)}$, $\mathbf{A}_g^{(t)}$, and $\mathbf{A}_g^{(a)}$ denote the spatial graph, temporal graph, and Self-adaptive graph, respectively.

The global predictor contains a closeness branch and a period branch, which are used to model recent dynamics and recurring patterns, respectively. For each input branch $r\in\{c,p\}$, FFT is first employed to extract frequency-domain features:
\begin{equation}
\mathbf{F}_g^r=\operatorname{FFT}(\mathbf{Y}^r).
\end{equation}
Then, the frequency-aware representation is fed into the temporal modeling process, where a pair of temporal convolution branches is used to capture temporal dependencies:
\begin{equation}
\mathbf{T}_g^r=
\tanh\!\big(\operatorname{TCN}_a(\mathbf{F}_g^r)\big)
\odot
\sigma\!\big(\operatorname{TCN}_b(\mathbf{F}_g^r)\big),
\end{equation}
where $\odot$ denotes element-wise multiplication.

In the structural modeling stage, graph convolutions are performed over the three inter-site graph views, yielding view-specific representations:
\begin{equation}
\mathbf{Z}_g^{r,(k)}
=
\operatorname{GCN}\!\left(\mathbf{A}_g^{(k)},\mathbf{T}_g^r\right),
\qquad
k\in\{s,t,a\}.
\end{equation}

To fully exploit the complementary information among different graph views, we introduce an Adaptive Multi-view Fusion (AMVF) module to adaptively reweight the graph representations from the three views. Specifically, the importance score of the $k$-th view is computed as
\begin{equation}
e_g^{r,(k)}=\phi_g\!\left(\mathbf{Z}_g^{r,(k)}\right),
\end{equation}
where $\phi_g(\cdot)$ denotes a learnable scoring function. The fusion weights are then normalized by a Softmax function:
\begin{equation}
\alpha_g^{r,(k)}
=
\frac{\exp\!\left(e_g^{r,(k)}\right)}
{\sum_{j\in\{s,t,a\}}\exp\!\left(e_g^{r,(j)}\right)}.
\end{equation}
Accordingly, the fused representation for branch $r$ is given by
\begin{equation}
\widetilde{\mathbf{Z}}_g^r
=
\sum_{k\in\{s,t,a\}}
\alpha_g^{r,(k)}\mathbf{Z}_g^{r,(k)}.
\end{equation}

After obtaining the fused representations from the closeness and period branches, they are merged and fed into the prediction head to generate the site-level forecasting result:
\begin{equation}
\widehat{\mathbf{Y}}^{t:t+H-1}
=
\operatorname{Head}_g\!\left(
\widetilde{\mathbf{Z}}_g^c+\widetilde{\mathbf{Z}}_g^p
\right).
\end{equation}
Here, $\widehat{\mathbf{Y}}^{t:t+H-1}\in\mathbb{R}^{M\times H\times C}$ denotes the site-level forecast over the next $H$ time steps. The forecast corresponding to each edge site is then returned to that site as global guidance.

\subsection{Local Stage: Edge-side STF Forecasting and Refinement}
After receiving the cloud-side forecast, each edge site performs local STF modeling at the VM level and fuses the local prediction with the returned global guidance to generate the final output.

For each site $m$, we construct an intra-site multi-view graph set $\mathcal{G}_m=[{\mathbf{A}_m^{(p)}, \mathbf{A}_m^{(l)}, \mathbf{A}_m^{(a)}}]$, where $\mathbf{A}_m^{(p)}$, $\mathbf{A}_m^{(l)}$, and $\mathbf{A}_m^{(a)}$ denote the physical graph, logical graph, and self-adaptive graph, respectively, which characterize correlations among VMs from different perspectives.

Similar to the global predictor, the local predictor also adopts a dual-input STF architecture. For each input branch $r\in\{c,p\}$, frequency-domain features are first extracted as
\begin{equation}
\mathbf{F}_m^r=\operatorname{FFT}(\mathbf{X}_m^r).
\end{equation}
Temporal dependencies are then modeled by gated temporal convolutions:
\begin{equation}
\mathbf{T}_m^r=
\tanh\!\big(\operatorname{TCN}_a(\mathbf{F}_m^r)\big)
\odot
\sigma\!\big(\operatorname{TCN}_b(\mathbf{F}_m^r)\big).
\end{equation}

Next, graph convolutions are conducted over the three intra-site graph views:
\begin{equation}
\mathbf{Z}_m^{r,(k)}
=
\operatorname{GCN}\!\left(\mathbf{A}_m^{(k)},\mathbf{T}_m^r\right),
\qquad
k\in\{p,l,a\}.
\end{equation}

Likewise, the local stage also employs AMVF to adaptively fuse the graph representations under different views:
\begin{equation}
e_m^{r,(k)}=\phi_m\!\left(\mathbf{Z}_m^{r,(k)}\right),
\end{equation}
\begin{equation}
\alpha_m^{r,(k)}
=
\frac{\exp\!\left(e_m^{r,(k)}\right)}
{\sum_{j\in\{p,l,a\}}\exp\!\left(e_m^{r,(j)}\right)},
\end{equation}
\begin{equation}
\widetilde{\mathbf{Z}}_m^r
=
\sum_{k\in\{p,l,a\}}
\alpha_m^{r,(k)}\mathbf{Z}_m^{r,(k)}.
\end{equation}

Accordingly, the local predictor outputs the VM-level local forecast:
\begin{equation}
\widehat{\mathbf{X}}_{m,\mathrm{local}}^{t:t+H-1}
=
\operatorname{Head}_l\!\left(
\widetilde{\mathbf{Z}}_m^c+\widetilde{\mathbf{Z}}_m^p
\right).
\end{equation}

Since the cloud-side forecast is generated at the site level, it is first mapped back to the VM level through a disaggregation layer, yielding the VM-aligned global guidance:
\begin{equation}
\widehat{\mathbf{X}}_{m,\mathrm{global}}^{t:t+H-1}
=
f_{\mathrm{dis}}\!\left(\widehat{\mathbf{Y}}_m^{t:t+H-1}\right).
\end{equation}

Finally, the VM-level forecast is obtained by adaptively fusing the local prediction and the disaggregated global guidance:
\begin{equation}
\widehat{\mathbf{X}}_m^{t:t+H-1}
=
\boldsymbol{\lambda}_m\odot \widehat{\mathbf{X}}_{m,\mathrm{local}}^{t:t+H-1}
+
(1-\boldsymbol{\lambda}_m)\odot \widehat{\mathbf{X}}_{m,\mathrm{global}}^{t:t+H-1},
\end{equation}
where $\boldsymbol{\lambda}_m$ is a learnable gating tensor.




\section{Experimental Setup}
\begin{table*}[t]
\caption{Prediction performance across multiple horizons on CPU dataset.}
\label{tab:cpu_results}
\scriptsize
\begin{tabular*}{\textwidth}{@{\extracolsep{-2pt}}l*{12}{c}}
\toprule
\multirow{2}{*}{Models} &
\multicolumn{3}{c}{H=1} &
\multicolumn{3}{c}{H=4} &
\multicolumn{3}{c}{H=8} &
\multicolumn{3}{c}{H=12} \\
\cmidrule(lr){2-4}\cmidrule(lr){5-7}\cmidrule(lr){8-10}\cmidrule(lr){11-13}
& MAE & SMAPE & $R^2$
& MAE & SMAPE & $R^2$
& MAE & SMAPE & $R^2$
& MAE & SMAPE & $R^2$ \\
\midrule

LSTM & 0.166 & 0.191 & 0.974 & 0.235 & 0.220 & 0.948 & 0.325 & 0.263 & 0.912 & 0.405 & 0.295 & 0.873 \\
Informer~\cite{zhou2021informer} & 0.514 & 0.559 & 0.830 & 0.488 & 0.523 & 0.845 & 0.489 & 0.509 & 0.851 & 0.499 & 0.507 & 0.848 \\
Autoformer~\cite{wu2021autoformer} & 0.387 & 0.800 & 0.910 & 0.403 & 0.758 & 0.901 & 0.445 & 0.759 & 0.885 & 0.483 & 0.772 & 0.870 \\
TimesNet~\cite{wu2023timesnet} & 0.206 & 0.214 & 0.962 & 0.236 & 0.219 & 0.948 & 0.280 & 0.235 & 0.934 & 0.319 & 0.251 & 0.921 \\
TimeXer~\cite{wang2024timexer} & 0.191 & 0.202 & 0.900 & 0.218 & 0.206 & 0.884 & 0.257 & 0.220 & 0.869 & 0.293 & 0.236 & 0.865 \\
iTransformer~\cite{liu2023itransformer} & 0.197 & 0.201 & 0.915 & 0.221 & 0.208 & 0.884 & 0.259 & 0.220 & 0.882 & 0.295 & 0.239 & 0.856 \\
Crossformer~\cite{zhang2023crossformer} & 0.202 & 0.206 & 0.936 & 0.233 & 0.212 & 0.929 & 0.274 & 0.224 & 0.917 & 0.315 & 0.242 & 0.897 \\

\midrule
STGCN & 0.179 & 0.320 & 0.974 & 0.242 & 0.329 & 0.949 & 0.327 & 0.355 & 0.914 & 0.406 & 0.382 & 0.877 \\
DCRNN & 0.222 & 0.243 & 0.953 & 0.269 & 0.244 & 0.936 & 0.352 & 0.270 & 0.903 & 0.430 & 0.299 & 0.865 \\
GWNET & 0.173 & 0.216 & 0.974 & 0.239 & 0.238 & 0.947 & 0.324 & 0.268 & 0.913 & 0.400 & 0.298 & 0.878 \\
ASTGCN~\cite{guo2019astgcn} & 0.414 & 0.375 & 0.887 & 0.423 & 0.360 & 0.878 & 0.462 & 0.366 & 0.858 & 0.507 & 0.383 & 0.832 \\
GCRN~\cite{seo2018gcrn} & 0.210 & 0.356 & 0.968 & 0.259 & 0.347 & 0.945 & 0.340 & 0.363 & 0.912 & 0.417 & 0.380 & 0.873 \\
HGCN~\cite{guo2021hgcn} & 0.212 & 0.456 & 0.971 & 0.259 & 0.437 & 0.948 & 0.330 & 0.453 & 0.918 & 0.389 & 0.473 & 0.889 \\
OGCGRN~\cite{guo2021ogcrnn} & 0.199 & 0.369 & 0.968 & 0.262 & 0.366 & 0.943 & 0.350 & 0.397 & 0.906 & 0.430 & 0.403 & 0.867 \\
OTSGCCN~\cite{guo2022otsggcn} & 0.163 & 0.237 & 0.975 & 0.232 & 0.251 & 0.948 & 0.318 & 0.279 & 0.914 & 0.397 & 0.305 & 0.877 \\

\midrule
ELASTIC~\cite{li2023elastic} & 0.163 & 0.189 & 0.976 & 0.213 & 0.202 & 0.959 & 0.265 & 0.221 & 0.940 & 0.307 & 0.236 & 0.920 \\
XELASTIC~\cite{10759304} & 0.159 & 0.188 & 0.976 & 0.202 & 0.200 & 0.961 & 0.247 & 0.218 & 0.948 & 0.279 & 0.233 & 0.937 \\
\textbf{DSTFView} & \textbf{0.147} & \textbf{0.141} & \textbf{0.979} & \textbf{0.192} & \textbf{0.163} & \textbf{0.962} & \textbf{0.232} & \textbf{0.180} & \textbf{0.949} & \textbf{0.262} & \textbf{0.191} & \textbf{0.940} \\
\bottomrule
\end{tabular*}
\end{table*}

\begin{table*}[t]
\caption{Performance comparison across multiple horizons on TP dataset.}
\label{tab:tp_results}
\scriptsize
\begin{tabular*}{\textwidth}{@{\extracolsep{-2.5pt}}l*{12}{c}}
\toprule
\multirow{2}{*}{Models} &
\multicolumn{3}{c}{H=1} &
\multicolumn{3}{c}{H=4} &
\multicolumn{3}{c}{H=8} &
\multicolumn{3}{c}{H=12} \\
\cmidrule(lr){2-4}\cmidrule(lr){5-7}\cmidrule(lr){8-10}\cmidrule(lr){11-13}
& MAE & SMAPE & $R^2$
& MAE & SMAPE & $R^2$
& MAE & SMAPE & $R^2$
& MAE & SMAPE & $R^2$ \\
\midrule

LSTM & 19.631 & 0.963 & 0.564 & 20.675 & 0.937 & 0.548 & 23.009 & 0.971 & 0.522 & 25.083 & 1.004 & 0.506 \\
Informer~\cite{zhou2021informer} & 18.256 & 1.328 & 0.779 & 17.870 & 1.222 & 0.773 & 18.234 & 1.205 & 0.777 & 18.698 & 1.219 & 0.771 \\
Autoformer~\cite{wu2021autoformer} & 12.380 & 1.451 & 0.903 & 12.714 & 1.385 & 0.887 & 14.008 & 1.379 & 0.871 & 15.232 & 1.383 & 0.854 \\
TimesNet~\cite{wu2023timesnet} & 5.764 & 0.335 & 0.976 & 7.018 & 0.359 & 0.950 & 8.822 & 0.383 & 0.934 & 10.326 & 0.417 & 0.918 \\
TimeXer~\cite{wang2024timexer} & 5.357 & 0.329 & 0.910 & 6.571 & 0.333 & 0.884 & 8.115 & 0.357 & 0.868 & 9.489 & 0.391 & 0.852 \\
iTransformer~\cite{liu2023itransformer} & 5.551 & 0.323 & 0.924 & 6.775 & 0.326 & 0.918 & 8.439 & 0.350 & 0.882 & 9.853 & 0.394 & 0.866 \\
Crossformer~\cite{zhang2023crossformer} & 5.695 & 0.326 & 0.958 & 6.879 & 0.330 & 0.922 & 8.723 & 0.374 & 0.905 & 10.127 & 0.408 & 0.899 \\

\midrule
STGCN & 6.394 & 1.375 & 0.986 & 8.618 & 1.329 & 0.940 & 12.002 & 1.333 & 0.884 & 14.905 & 1.367 & 0.848 \\
DCRNN & 33.562 & 0.514 & 0.075 & 32.616 & 0.568 & 0.119 & 34.300 & 0.612 & 0.123 & 35.634 & 0.826 & 0.127 \\
GWNET & 4.889 & 0.710 & 0.981 & 7.123 & 0.684 & 0.945 & 10.387 & 0.718 & 0.899 & 13.281 & 0.762 & 0.853 \\
ASTGCN~\cite{guo2019astgcn} & 13.726 & 1.167 & 0.819 & 13.760 & 1.091 & 0.813 & 14.784 & 1.095 & 0.817 & 15.958 & 1.109 & 0.801 \\
GCRN~\cite{seo2018gcrn} & 15.820 & 1.181 & 0.733 & 16.804 & 1.125 & 0.726 & 18.948 & 1.159 & 0.700 & 21.022 & 1.183 & 0.674 \\

HGCN~\cite{guo2021hgcn} & 6.427 & 0.929 & 0.980 & 7.761 & 0.903 & 0.944 & 10.075 & 0.937 & 0.918 & 12.199 & 0.960 & 0.882 \\
OGCGRN~\cite{guo2021ogcrnn} & 16.981 & 1.222 & 0.734 & 18.185 & 1.186 & 0.718 & 20.559 & 1.200 & 0.682 & 22.723 & 1.234 & 0.656 \\
OTSGCCN~\cite{guo2022otsggcn} & 3.845 & 1.156 & 0.987 & 6.539 & 1.100 & 0.951 & 9.903 & 1.124 & 0.895 & 12.726 & 1.098 & 0.869 \\

\midrule
ELASTIC~\cite{li2023elastic} & 3.859 & 0.290 & 0.981 & 5.633 & 0.314 & 0.965 & 7.706 & 0.348 & 0.949 & 9.250 & 0.382 & 0.933 \\
XELASTIC~\cite{10759304} & 3.782 & 0.314 & 0.995 & 5.446 & 0.318 & 0.973 & 7.150 & 0.352 & 0.953 & 8.374 & 0.385 & 0.937 \\
\textbf{DSTFView} & \textbf{3.496} & \textbf{0.237} & \textbf{0.998} & \textbf{5.189} & \textbf{0.251} & \textbf{0.979} & \textbf{6.753} & \textbf{0.295} & \textbf{0.957} & \textbf{7.877} & \textbf{0.319} & \textbf{0.939} \\
\bottomrule
\end{tabular*}
\end{table*}

\subsection{Datasets}
To evaluate the effectiveness of \emph{DSTFView}, we conduct experiments on two real-world datasets collected from Alibaba Cloud-native Edge Node Service (ENS) and the Network Edge Platform (NEP)\footnote{\url{https://github.com/xumengwei/EdgeWorkloadsTraces}.}. These datasets represent typical workload patterns in large-scale cloud--edge environments.

\textbf{CPU dataset}, collected from Alibaba ENS, it contains CPU utilization traces from 1,728 IaaS virtual machines (VMs) across 77 geographically distributed edge sites in China. 
\textbf{TP dataset}, collected from Alibaba NEP, it records average network traffic, including uplink and downlink throughput measured in $Mbps$. Compared with CPU utilization, network throughput is more sensitive to user mobility and bursty service requests.



\subsection{Experimental Details}
We adopt Z-score standardization and divide the dataset into training, validation, and test subsets in chronological order with a ratio of 7:2:1. All proposed methods are trained with the Adam optimizer, using an initial learning rate of 0.001 and an ExponentialLR scheduler for decay. The loss function is mean squared error (MSE). For both the global and local stages, the batch size is set to 64 and the maximum number of training epochs is 50.
All experiments are executed on a Linux server equipped with four NVIDIA TITAN X (Pascal) GPUs, each with 12\,GB memory.

\section{Experimental Results}
\begin{table*}[t]
\caption{Ablation study of DSTFView under different global graph configurations.}
\label{tab:global_ablation}
\scriptsize
\begin{tabular*}{\textwidth}{@{\extracolsep{-4pt}}ll*{8}{c}}
\toprule
\multirow{2}{*}{Datasets} & \ \ \multirow{2}{*}{Configurations} &
\multicolumn{2}{c}{H=1} &
\multicolumn{2}{c}{H=4} &
\multicolumn{2}{c}{H=8} &
\multicolumn{2}{c}{H=12} \\
\cmidrule(lr){3-4}\cmidrule(lr){5-6}\cmidrule(lr){7-8}\cmidrule(lr){9-10}
& & MAE & SMAPE
  & MAE & SMAPE
  & MAE & SMAPE
  & MAE & SMAPE \\
\midrule

\multirow{7}{*}{CPU}
& Identity                    & 2.953 & 0.050 & 3.965 & 0.076 & 5.208 & 0.111 & 6.552 & 0.145 \\
& spatial-only                & 2.557 & 0.050 & 3.241 & 0.068 & 4.174 & 0.091 & 5.203 & 0.114 \\
& temporal-only               & 2.745 & 0.050 & 3.579 & 0.072 & 4.599 & 0.099 & 5.699 & 0.126 \\
& adaptive-only               & 2.409 & 0.050 & 2.932 & 0.064 & 3.721 & 0.082 & 4.608 & 0.100 \\
& spatial-temporal            & 2.379 & 0.049 & 2.894 & 0.064 & 3.664 & 0.081 & 4.489 & 0.098 \\
& \textbf{tim-spa-adaptive(AMVF)} & \textbf{2.348} & \textbf{0.049} & \textbf{2.826} & \textbf{0.063} & \textbf{3.579} & \textbf{0.079} & \textbf{4.370} & \textbf{0.095} \\

\midrule

\multirow{7}{*}{TP}
& Identity                    & 703.272 & 0.307 & 771.352 & 0.325 & 932.822 & 0.367 & 1152.779 & 0.416 \\
& spatial-only                & 622.728 & 0.280 & 683.011 & 0.297 & 825.989 & 0.335 & 1020.755 & 0.379 \\
& temporal-only               & 391.445 & 0.203 & 429.338 & 0.215 & 519.213 & 0.242 & 641.642 & 0.274 \\
& adaptive-only               & 556.219 & 0.258 & 610.064 & 0.273 & 737.771 & 0.308 & 911.735 & 0.349 \\
& spatial-temporal            & 225.359 & 0.147 & 247.175 & 0.156 & 298.917 & 0.176 & 369.401 & 0.199 \\
& \textbf{tim-spa-adaptive(AMVF)} & \textbf{214.357} & \textbf{0.143} & \textbf{235.108} & \textbf{0.152} & \textbf{284.324} & \textbf{0.171} & \textbf{351.367} & \textbf{0.194} \\

\bottomrule
\end{tabular*}
\end{table*}

\begin{table*}[t]
\caption{Ablation study of DSTFView under different local graph configurations.}
\label{tab:local_graphs_amvf}
\scriptsize
\begin{tabular*}{\textwidth}{@{\extracolsep{-0.5pt}}ll*{8}{c}}
\toprule
\multirow{2}{*}{Datasets} & \multirow{2}{*}{Configurations} &
\multicolumn{2}{c}{H=1} &
\multicolumn{2}{c}{H=4} &
\multicolumn{2}{c}{H=8} &
\multicolumn{2}{c}{H=12} \\
\cmidrule(lr){3-4}\cmidrule(lr){5-6}\cmidrule(lr){7-8}\cmidrule(lr){9-10}
& & MAE & SMAPE
  & MAE & SMAPE
  & MAE & SMAPE
  & MAE & SMAPE \\
\midrule

\multirow{5}{*}{CPU}
& Identity                    & 0.182 & 0.157 & 0.238 & 0.181 & 0.288 & 0.200 & 0.324 & 0.213 \\
& physical-only               & 0.959 & 0.362 & 1.253 & 0.416 & 1.518 & 0.461 & 1.711 & 0.489 \\
& logical-only                & 0.168 & 0.151 & 0.219 & 0.174 & 0.265 & 0.193 & 0.299 & 0.204 \\
& adaptive-only               & 0.154 & 0.145 & 0.201 & 0.167 & 0.243 & 0.184 & 0.274 & 0.196 \\
& \textbf{phy-log-adaptive (AMVF)} & \textbf{0.147} & \textbf{0.141} & \textbf{0.192} & \textbf{0.163} & \textbf{0.232} & \textbf{0.180} & \textbf{0.262} & \textbf{0.191} \\

\midrule

\multirow{5}{*}{TP}
& Identity                    & 4.363 & 0.265 & 6.478 & 0.280 & 8.430 & 0.329 & 9.833 & 0.356 \\
& physical-only               & 3.805 & 0.247 & 5.649 & 0.262 & 7.351 & 0.308 & 8.575 & 0.332 \\
& logical-only                & 3.817 & 0.248 & 5.667 & 0.262 & 7.375 & 0.308 & 8.602 & 0.333 \\
& adaptive-only               & 3.641 & 0.242 & 5.406 & 0.256 & 7.035 & 0.301 & 8.206 & 0.325 \\
& \textbf{phy-log-adaptive (AMVF)} & \textbf{3.496} & \textbf{0.237} & \textbf{5.189} & \textbf{0.251} & \textbf{6.753} & \textbf{0.295} & \textbf{7.877} & \textbf{0.319} \\
\bottomrule
\end{tabular*}
\end{table*}
\subsection{Overall Comparison}
Tables~\ref{tab:cpu_results} and~\ref{tab:tp_results} report the forecasting results on the CPU and TP datasets over four prediction horizons (1/4/8/12) in terms of MAE, SMAPE, and $R^2$. We compare \emph{DSTFView} with representative temporal, spatiotemporal, and collaborative cloud--edge baselines.

Temporal baselines, such as iTransformer and TimerXer, model each site or VM independently and therefore cannot explicitly capture inter-entity dependencies. Spatiotemporal methods, such as OGCGRN and OTSGCCN, consider graph relations, but many rely on a single predefined graph and may not fully reflect the heterogeneous correlations in edge workloads. ELASTIC and XELASTIC adopt a collaborative cloud--edge pipeline, but they cannot jointly model temporal and frequency patterns or adaptively exploit multiple relational views.

As shown in Tables~\ref{tab:cpu_results} and~\ref{tab:tp_results}, \emph{DSTFView} consistently achieves the best performance on both datasets. This advantage mainly comes from two aspects. Firstly, the proposed dual-input STF modeling captures recent dynamics, periodic patterns, and spatial correlations in a unified manner. Also, the adaptive multi-view fusion mechanism dynamically integrates multiple relational views in both the global and local stages, thereby enabling more effective dependency modeling and improving forecasting accuracy.

\subsection{Ablation Study}
In this section, we conduct an ablation study to evaluate the contributions of its key components at the \emph{Global Stage} and the \emph{Local Stage}.

For the global stage, we examine single-view variants, a dual-view variant (\emph{spatial-temporal}), and the full model integrating temporal, adaptive, and spatial views. As reported in Table~\ref{tab:global_ablation}, multi-view combinations generally outperform single-view settings, and the complete model achieves the best overall results, demonstrating the benefit of jointly modeling temporal, frequency, and spatial information.

For the local stage, we compare different relation configurations, including Identity, physical-only, logical-only, adaptive-only, and the fused \emph{phy-log-adaptive} setting (AMVF). Table~\ref{tab:local_graphs_amvf} shows that adaptive relations consistently improve over the baseline and other single-view settings, while the fused AMVF configuration achieves the best performance, confirming the effectiveness of combining multiple relation views.
\subsection{Visualization Analysis}
\begin{figure}[t]
  \centering
  \includegraphics[width=0.5\columnwidth]{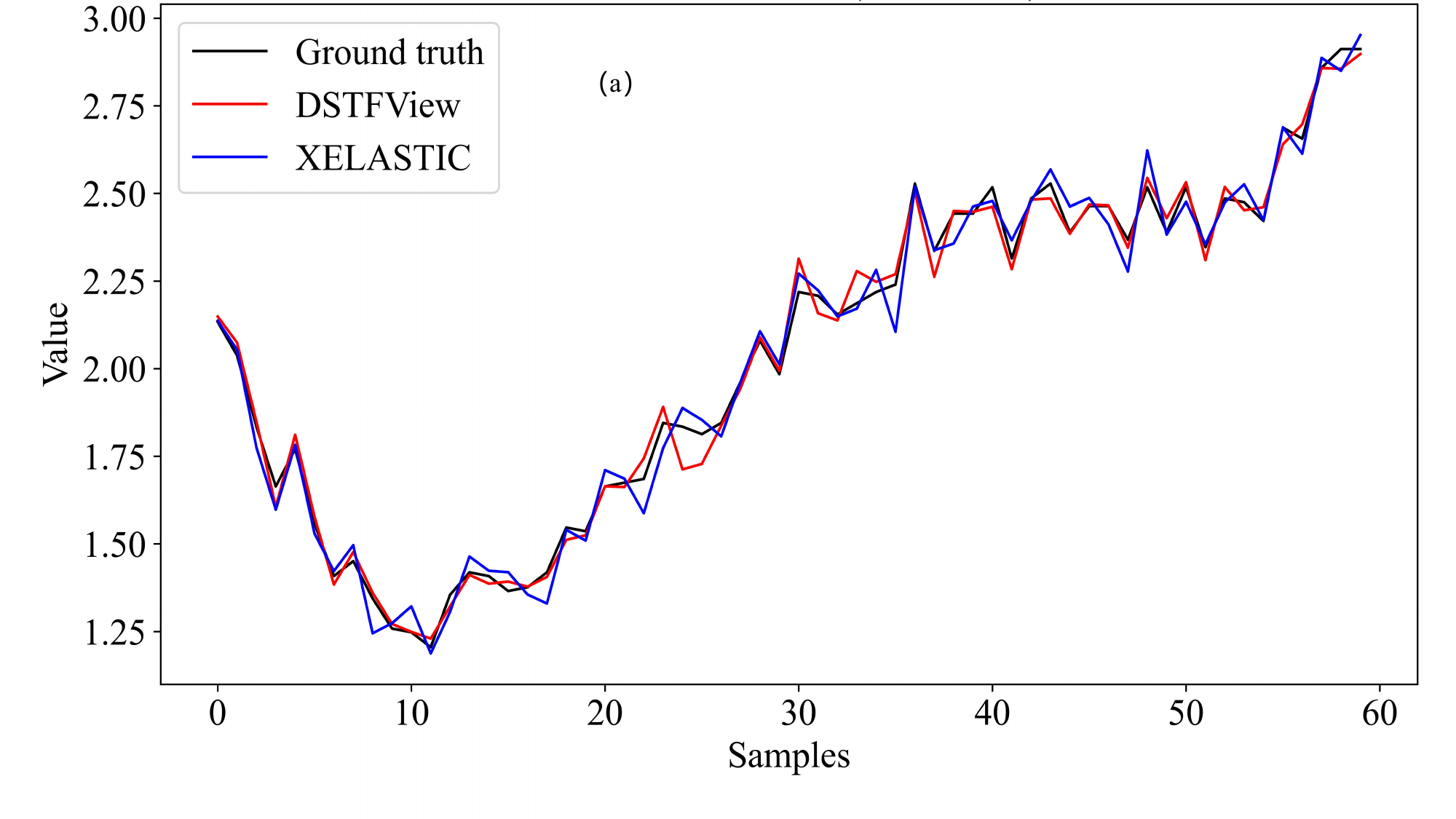}\hfill
  \includegraphics[width=0.5\columnwidth]{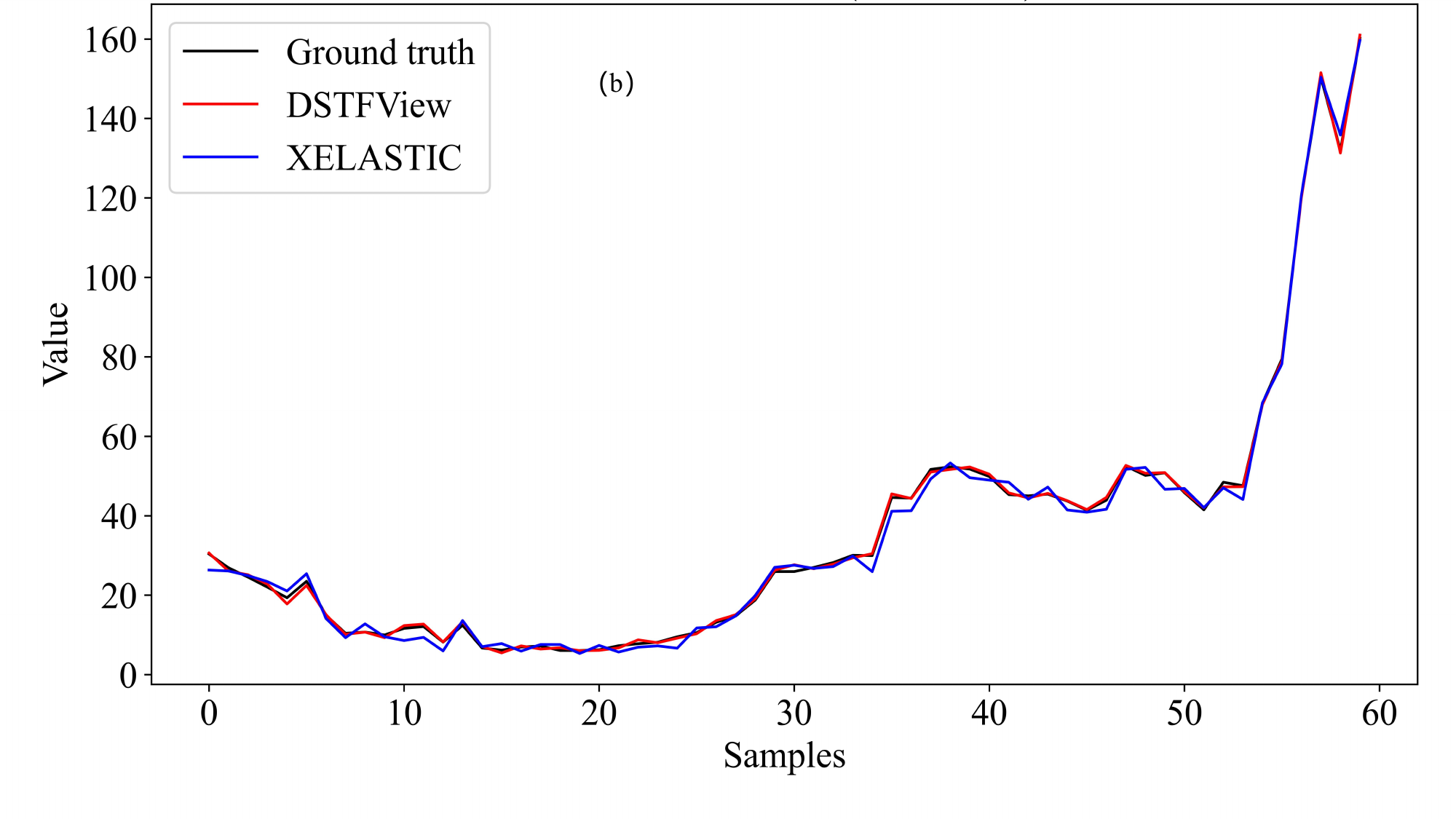}
  \caption{Visualization results on the CPU and TP dataset.}
  \label{fig2:visualization}
\end{figure}
Fig.~\ref{fig2:visualization} presents qualitative forecasting examples for a representative VM under the one-step-ahead setting, where subfigure (a) corresponds to the CPU dataset and subfigure (b) corresponds to the TP dataset. In the figure, the black curve indicates the ground-truth series, the red curve denotes \emph{DSTFView}, and the blue curve represents the strongest competing baseline, \emph{XELASTIC}. As can be observed from Fig.~\ref{fig2:visualization}, \emph{DSTFView} tracks the target series more closely, especially around peaks, valleys, and local trend reversals, whereas \emph{XELASTIC} exhibits larger deviations in these regions. This advantage can be attributed to the proposed model's ability to jointly capture periodic components and short-term high-frequency fluctuations through adaptive AMVF-based fusion, thereby yielding more accurate predictions on both datasets.

\section{Conclusion}
In this study, we propose \emph{DSTFView}, a collaborative cloud--edge framework for workload forecasting. It uses dual-input spatio-temporal-frequency multi-view modeling to jointly capture cross-site dependencies, recent dynamics, and periodic patterns through cloud-side global forecasting and edge-side local forecasting. The adaptive fusion mechanism further highlights informative graph structures.  Experimental results on the CPU and TP datasets demonstrate that  \emph{DSTFView} consistently outperforms representative baselines across multiple forecasting horizons. Future work will focus on dynamic graph construction, continual learning, and integration with resource management tasks such as auto-scaling and scheduling to improve adaptability and practical value in real edge environments.


\bibliographystyle{splncs04}
\bibliography{references}
\end{document}